\documentclass[letterpaper]{article}
\usepackage{tmp}
\usepackage{times}
\usepackage{helvet}
\usepackage{courier}
\usepackage{graphicx}
\usepackage{todonotes}
 
\usepackage{graphicx}
\usepackage{caption}
\usepackage{subcaption}
\usepackage{amsmath}
\usepackage{xcolor}
\usepackage{color, soul}

\usepackage{booktabs}
\usepackage{multirow}
\newcommand{\gray}[1]{\textcolor{gray}{#1}}
\newcommand{\red}[1]{\textcolor{red}{#1}}
\definecolor{cobalt}{rgb}{0.0, 0.28, 0.67}
\newcommand{\blue}[1]{\textcolor{cobalt}{#1}}
\begin{document}

\title{Explaining Link Predictions in Knowledge Graph Embedding Models with Influential Examples}
\author{Adrianna Janik \textsuperscript{1}, Luca Costabello \textsuperscript{1}\\
\textsuperscript{1} Accenture Labs}
\maketitle
\begin{abstract}
We study the problem of explaining link predictions in the Knowledge Graph Embedding (KGE) models. We propose an example-based approach that exploits the latent space representation of nodes and edges in a knowledge graph to explain predictions. We evaluated the importance of identified triples by observing progressing degradation of model performance upon influential triples removal. Our experiments demonstrate that this approach to generate explanations outperforms baselines on KGE models for two publicly available datasets.
\end{abstract}

\section{Introduction}\label{sec:introduction}
Link prediction is a common task in data represented as knowledge graphs. One family of models: graph embedding models are used across different domains to tackle link prediction tasks in a parameter-efficient manner. There is an increasing legal requirement for any machine learning model utilised in the industry to provide explanations of model predictions. Despite its importance, there is a lack of explainability methods for knowledge graph embedding models \cite{bianchi_knowledge_2020}. Few works that exist \cite{lawrence_explaining_2020,bhardwaj_poisoning_2021,kang_explaine_2019,betz_adversarial_2022} differ in evaluation approaches and datasets used for the evaluation. Human readability of explanations is also an unexplored topic among the existing works and usually does not go beyond providing a few anecdotal cases drawn from the dataset. In this work, we propose a new method of explaining predictions of knowledge graph embedding models with influential examples and propose a dataset to evaluate explanations in user studies.

There is an increasing legal need for AI systems to be explainable. On the 21st of April, 2021 EU Commission proposed the first-ever legal framework regulating AI \cite{noauthor_eur-lex_nodate}. When this act will be enforced high-risk AI such as the one used in the medical systems should “facilitate the interpretation of the outputs of AI systems by the users” (Article 13) adding explanation requirements on top of the rights specified in GDPR.

\begin{figure}[hbt!]
    \centering
    \includegraphics[scale=0.40]{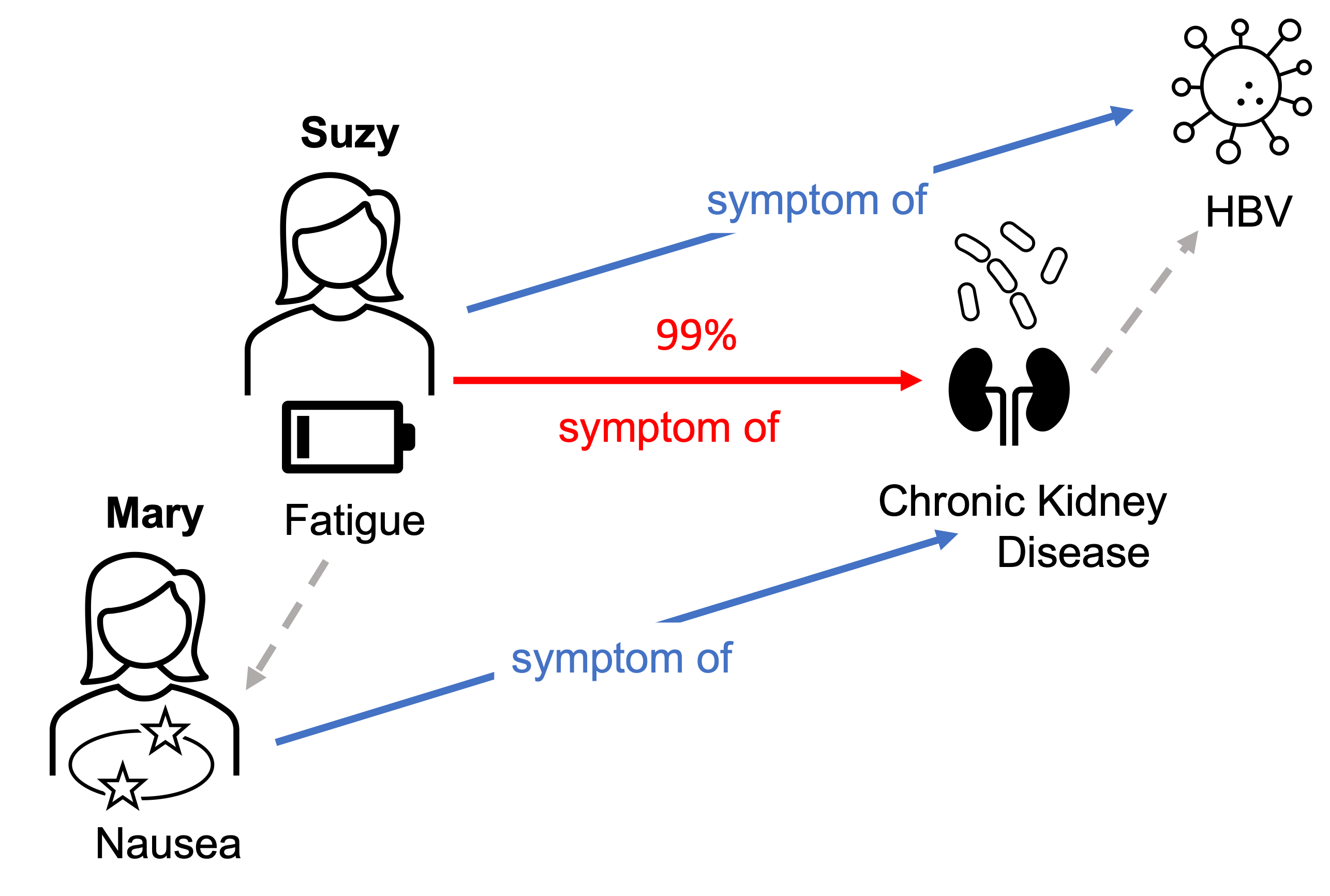}
    \caption{To support \red{prediction of the target statement} we identify \blue{influential examples} by probing the knowledge base \gray{constrained w.r.t. the latent-space}. This example is drawn from the Fb15k-237 dataset. Predicted plausability score was 99\%, and two most influential examples were retrieved as an explanation with the following ranks: 1st: $Nausea \xrightarrow{} symptomOf \xrightarrow{} Chronic Kidney Disease$, 2nd: $Fatigue \xrightarrow{} symptomOf \xrightarrow{} HBV$}.
    \label{fig:my_label}
\end{figure}

Link prediction conducted with Knowledge Graph Embeddings (KGE) models like ComplEx, DistMult, TransE etc... despite being accurate suffers from not being directly interpretable \cite{bianchi_knowledge_2020}. The user usually as a result of prediction receives a single number corresponding to a score that further needs to be interpreted or the model needs to be calibrated \cite{tabacof_probability_2020} in such a way that it returns the estimated probability score. As KGEs models like link prediction models are being applied in critical tasks \cite{kge_tutorial} e.g. drug discovery and medicine where low interpretability can have serious consequences, it is of paramount importance to develop effective interpretability methods for this kind of models. At the moment, there is no way to directly understand what contributed to a prediction of a KGE model. One way to tackle this issue is to use special interpretability methods that work post-hoc. Another way is to design new approaches inspired by the methods available for explaining other machine learning models \cite{guidotti_survey_2018}. An example of desirable output would be such an explanation that links the prediction back to the original graph pointing out to links and nodes (and other attributes) that contributed the most to the given prediction and which removal would result in a decreased probability of the prediction, being in the same time understandable to users. It is also desirable that we can obtain the explanations fast and that they are also memory efficient. Given that, we focused on the following research question:

\subsubsection{Research Question} How to provide pertinent explanations for relational learning models trained on large knowledge graphs with reasonable time/memory constraints?

\subsubsection{Explainability Challenges}\label{sec:explainability_challenges}

When addressing the problem of explainability of the machine learning models we discover new problems. Starting from the definition of what explainability is in the context of knowledge graph embeddings, through lack of evaluation protocols for the methods, lack of metrics, XAI benchmark datasets that would not only facilitate experiments on underlying predictive task like link prediction but also explainability. Common benchmark dataset exists to rank models based on their performance FB15k-237, WN18RR, YAGO3\_10, and others these however were not build to assess their explainability. Often these datasets suffer from lack of human-readable labels, e.g. in FB15k-237 the ids are encoded and the dataset does not come with the mapper file for the entities and relations. Given that Freebase dataset was discontinued it is becoming obsolete very quickly and retrieval of labels becomes a tedious task. We can easily recall an example from popular benchmark Fb15k-237 of a triple which is clearly not readable: /m/08966, /travel/travel\_destination/climate./travel/travel\_destination\_-monthly\_climate/month, /m/05lf\_. Not only ids of entities are unknown but also predicates structure is cumbersome and not easily understandable. Similar situation happens also in the case of WN18RR dataset: 06845599, \_member\_of\_domain\_usage, 03754979. Despite their lack of interpretability in  the sense of understanding what each triple means these benchmark datasets are useful and help establish common grounds for the community. Some newer datasets are more interpretable but not as widely used as others, e.g. CoDEx \cite{safavi_codex_2020} comes not only with understandable labels, descriptions and sources but also with multiple languages and comes from the data source that is actively in usage (Wikidata and Wikipedia). Next there is a lack of ground-truth explanations which brings us back to the problem of the definition of the explanations. Finally explainability approaches are often treated separately from their actual interpretability by potential users. Very few works present the user studies on the explanations.

\subsubsection{Contribution}\label{sec:contribution}
In this work we present:
\begin{itemize}
    \item  ExamplE heuristics to generate explanation graphs and time-efficient batch mode for generating influential examples.
    \item Evaluation protocol involving novel XAI test set for evaluating interpretability of explanations for the users, based on Fb15k-237 benchmark dataset.
    \item Code along with the dataset and generated explanations will be available as a part of AmpliGraph Python library.
\end{itemize}

\subsection{Explaining Link Predictions with Influential Examples and Explanation Graph}

\begin{table}[hbt!]
\begin{tabular}{l|llll}
           & Fb15k-237 & WN18RR  & XAI-Fb15k-237 \\
\hline
Train   &    272,115       &   86,835   &  - \\
Test       &    20,466       &  3,134    & 239 \\
Valid &    17,535       &  3,034    &  - \\
\hline
Entities &     14,541      &    40,943   & 445    \\
Relations &     237      &      11  &  91 \\
\end{tabular}
\caption{Details of the two benchmark datasets Fb15k-237 and WN18RR utilised to evaluate performance of the explainability approach and a novel explainability testset prepared within the scope of this work: XAI-Fb15k-237.}
\end{table}

In this section we introduce the concept of explanation graph. We then move to the steps required to obtain explanation graph and speed-up approach to obtain influential examples. Next we present the evaluation protocol employed in this work and metrics for measuring the performance of the explainability approach presented. As an addition we will present an introduction to a small test set we have prepared for testing XAI methods with the goal of reusing it for future user studies.

Intuition. We propose ExamplE, a post-hoc, local explanation subsystem that explains predictions
of links returned by any knowledge graph embedding model architecture. Our novel ExamplE
method is based on the assumption that to explain why a certain link between two entities is
predicted as plausible, both the n-hop neighbourhoods of the specific entities and predicate should
be considered as well as how these are related to the training samples that the pattern was
extracted from (i.e. which training examples were yielded as similar - influential examples). As a result, the returned
explanation includes both important examples buried in the latent space and a subgraph of the
training graph rooted at the target triple being predicted and explained. The combination of these
two components, equally important, to create a new form of explanation has not been adopted by
prior art.

\begin{figure*}[ht!]
    \centering
    \includegraphics[width=\textwidth]{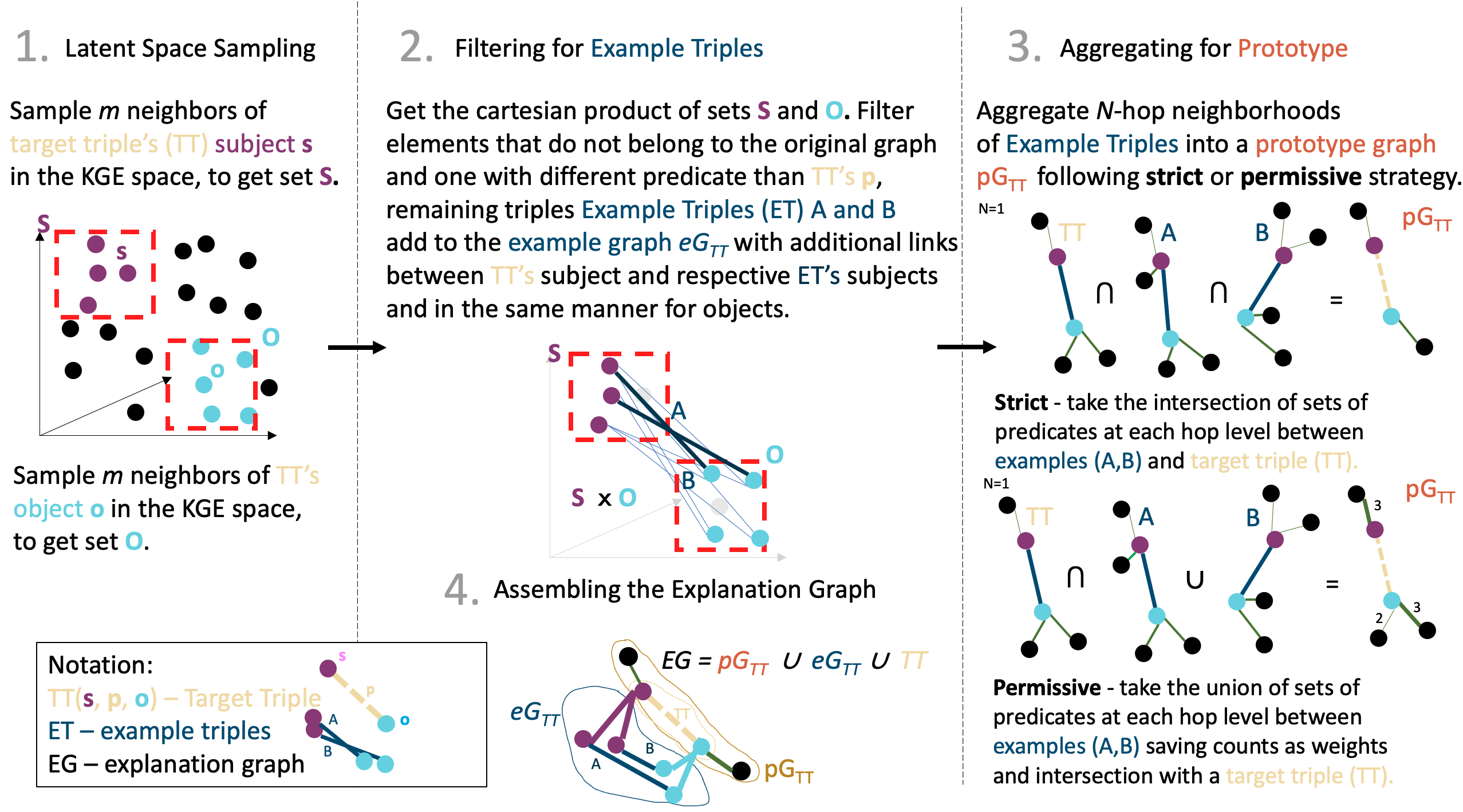}
    \caption{Detailed steps of the example-based explanation heuristic. Proposed method consists of four steps: Latent Space Sampling, Filtering for Example Triples, Aggregating for Prototype, and Assembling the Explanation Graph.}
    \label{fig:diagram}
\end{figure*}

There are four main steps that are core to this approach, described in detail in the following section.

\subsubsection{Prerequisites:}

    Calibrated KGE model, returning probability estimates (bounded scores at the output).

\subsubsection{The Explanation Graph:} is composed of weighted n-hop
neighbourhoods (up to a certain hop-level n) where weights are assigned based on relevance. This
makes up the Target Triple Neighbourhood Plane (Figure 11). Examples that are closely related to
the target triple and have similar embeddings as the target triple are included in the Examples Plane
(Figure 11). Example triples from the Examples Plane are connected with the target triple with new
links that indicate resemblance between the target triple object and the objects of triples in the
Example Plane (likewise for triple subjects, see Figure 11).
\begin{figure}
    \centering
    \includegraphics[width=0.5\textwidth]{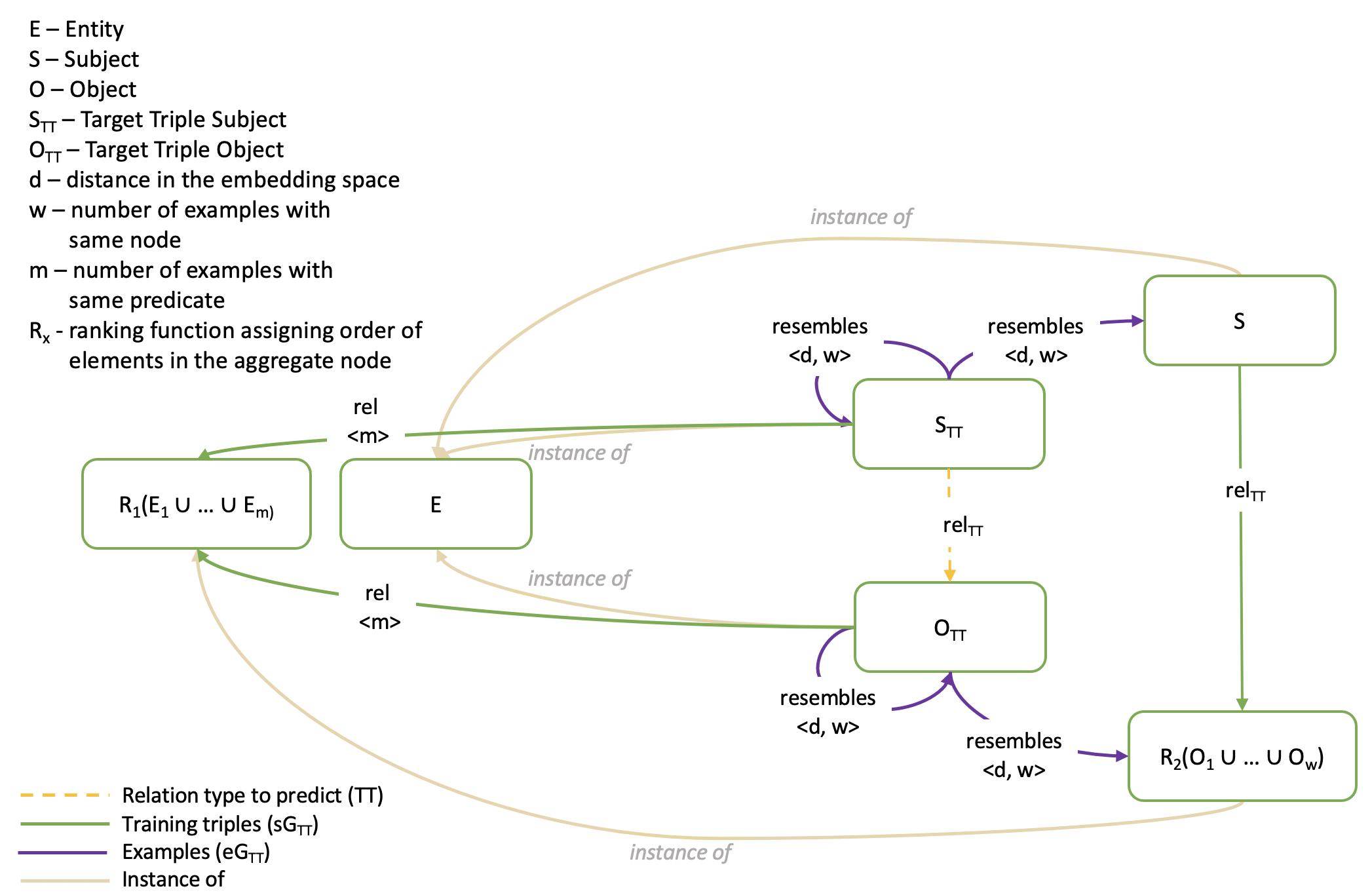}
    \caption{Explanation graph format.}
    \label{fig:my_label}
\end{figure}

\subsubsection{Method:} ExamplE is an example-based heuristics that consists of four steps: sampling, filtering for examples, aggregating for prototype and assembling the Explanation Graph. The steps are
described below and are summarised in Figure 2.
\begin{enumerate}
\item Latent space sampling: samples m neighbours of target triple's (TT) subject s in the KGE
space, to get a set S. Sample m neighbors of TT’s object o in the KGE space, to get set O. For each neighbour an associated score is saved that defines how similar the neighbour was to the respective element.
\item Filter for example triples: Get the cartesian product of sets S and O. Filter elements that
do not belong to the original graph and one with different predicate than TT’s predicate.
Remaining triples are called Example Triples ($ET$) $A$ and $B$ (Figure 2) and are added to the example graph $eGTT$ (Figure 2) with
additional links between TT’s subject and respective $ET$’s subjects and in the same manner
for objects. In this step also a joint score for the example triple is calculated, by taking a weighted average of scores of all respective elements.
\item Aggregating for Prototype: Aggregate N-hop neighborhoods of Example Triples into a
prototype graph $pGTT$ following strict or permissive strategy.
1. Strict - take the intersection of sets of predicates at each hop level between
examples ($A$,$B$) and target triple (TT).
2. Permissive - take the union of sets of predicates at each hop level between
examples ($A$,$B$) saving counts as weights and intersection with a target triple (TT).
\item Assemble the Explanation Graph: the last step combines the results of the previous steps
into an Explanation Graph.
\end{enumerate}

Above spatial constraints let to fetch all of examples within specified constraints by the parameters from the Target Triple upon availability which means that None can also be returned as a valid output if there are no similar examples within that constraints.

\section{Evaluation}\label{sec:evaluation}
To evaluate the method we generated explanations for TransE model on two different datasets with our proposed approach and also using a baseline approach and then modified the training dataset based on the explanations and retrained the models. To evaluate our explanation approach we consider only generation of influential examples with associated scores and treating explanation graph as an extra add-on to analyze further the explanations if required.

\subsubsection{Baselines:}
As a baseline we utilized random explanation approach where we limit triples only to the ones with same predicate without any constrains on how and if such triple is connected to the target triple. Some other baselines could have been utilised like random triple from the neighbourhood of the target triple, this baseline however made the task more complicated since in the scenario we look at explanations with the same predicates, making the whole space of available triples for explanations limited to this subset.

\subsubsection{Hypothesis:} Retrained model scores target triple as less plausible (compared to the original model scoring the same target triple) upon removal of the explanation (influential examples identified by the explainer).
To test this hypothesis we first had to identify a candidate for the target triple. Initially we assumed that highest scoring triple on the test set is a good candidate for the target triple.
General Procedure: Firstly, predicted probabilities were collected for the whole test dataset. Secondly, test triples were ranked based on the probabilities. Triple with the highest probability predicted was selected for the experiment, we call it a Target Triple (TT). After initial experiments it turned out that the highest scoring triples were circular (subject equals to object), therefore we narrowed down selection to first-non-circular highest scoring triple. We obtained explanation for the Target Triple with ExamplE, using the default parameters. We finally modified the dataset in different scenarios guided by the obtained explanation to assess difference in predicted probabilities for the Target Triple for the original model and model trained on the modified dataset.

\subsubsection{Scenarios:}
\begin{enumerate}
    \item{Remove-and-Retrain (ROAR) \cite{hooker_benchmark_2019}: We removed explanation from the dataset and retrained the model (ROAR protocol) on the modified dataset without explanation for two cases removing only the most influential example triple and removing full set of examples returned by the method.}
    \item{Reversed-Remove-and-Retrain (rev-ROAR): We removed all triples with same predicate as Target Triple and instead added only the explanation (influential examples are restricted to be of the same predicate type as a target triple by default). In this scenario we wanted to test whether model can recover from a loss of majority of its influential examples. We also explored two cases leaving only the most influential example triple and leaving the full set of examples returned by the method.}
\end{enumerate}    

Note: Both models: original and the one retreined in each experiment were trained exactly the same, with SOTA hyperparameters for the TransE model on the dataset with early stopping. This has the following consequence of evaluation time being very long, for the four scenarios presented we had to train 5 models per dataset per target triple.

\subsubsection{Implementation:}
All experiments were implemented using Python 3.7 with Knowledge
Graph Embedding library AmpliGraph version 2.0, using TensorFlow 2.10 . All experiments
were run under Ubuntu 18.04 on an Intel Xeon Gold 6142, 64 GB, equipped with a Tesla V100
16GB.

\subsection{Results}\label{sec:experiments}

In this section we present the experiments we have run to test the following hypothesis:
\begin{enumerate}
    \item Hypothesis: Probabilities are highly correlated before and after retraining and the slope coefficient is $< 1$, meaning that the original probabilities are higher than respective probabilities after retraining.

    \item Hypothesis: Plausibility of triples are the more lower the more explaining triples are removed with magnitude of a drop reflecting the rank of triples importance.
    
\end{enumerate}

Both evaluation scenarios are reported in figures: 3-8 and tables 2-3.
The first hypothesis was confirmed in all the cases for our method and random, scores after retraining were correlated and whenever we removed the explaining triples with same predicate the respective plausibility after retraining was lower, see Figures 5 for Fb15k-237 and 6 for WN18RR.
We can see how effectiveness of the explanations depends on how long the model had to recover since the longer the training time the lower the difference between probabilities with original model and the one with influential triples removed. In the case of Fb15k-237 we could see that model recovered almost fully to predicting triple as plausible after 80 epochs of training (Figure 8). This was not the case for WN18RR (Figure 7) where single triple could not convey this much information as in the previous dataset.
As for the second hypothesis we could see that the more triples were removed from training the bigger the difference between probabilities become, but what was interesting was that the longer the model was trained the differences become smaller, suggesting that the model could somehow recover from triples removal and use other triples remaining in the dataset to deduce the plausability. For three datasets we present examples of generated explanations in Tables 10-12. We also recall times of explanation generation for models trained on different datasets in Table 5.

\begin{figure}
    \centering
    \includegraphics[width=0.5\textwidth]{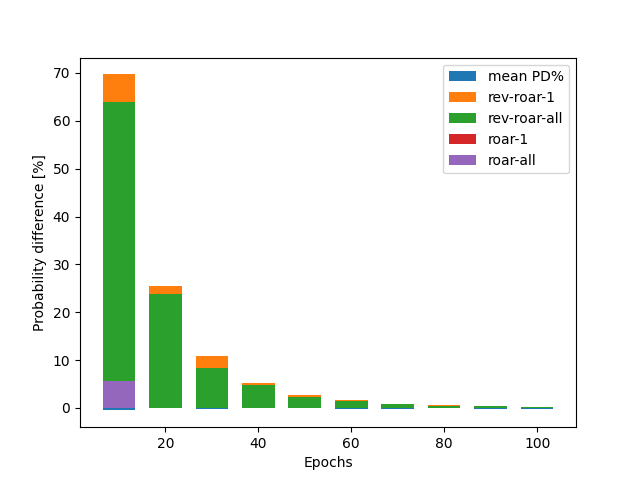}
    \caption{TransE on Fb15k-237 - Probability difference (PD) between test triples before and after retraining, average PD for all triples and PD of the target triple per epoch. Under rev-ROAR scenario.}
    \label{fig:my_label}
\end{figure}

\begin{figure}
    \centering
    \begin{subfigure}[b]{0.23\textwidth}
    \includegraphics[width=\textwidth]{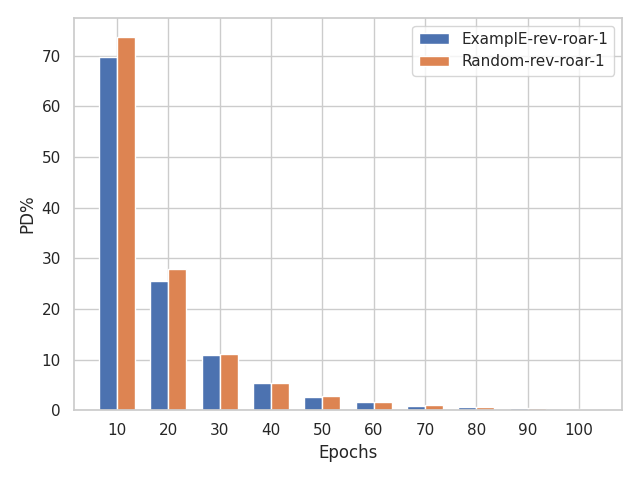}
    \end{subfigure}
    \hfill
    \begin{subfigure}[b]{0.23\textwidth}
    \includegraphics[width=\textwidth]{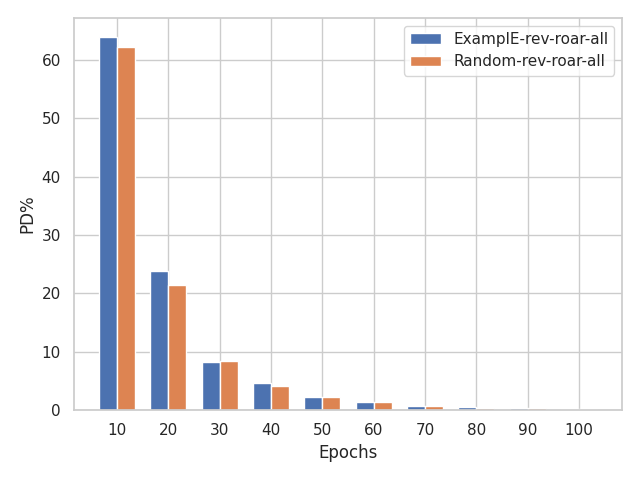}
    \end{subfigure}    
        \hfill
    \begin{subfigure}[b]{0.23\textwidth}
    \includegraphics[width=\textwidth]{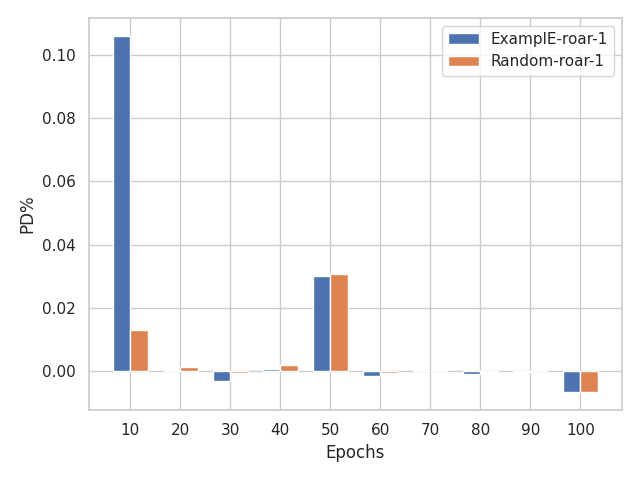}
    \end{subfigure}
    \begin{subfigure}[b]{0.23\textwidth}
    \includegraphics[width=\textwidth]{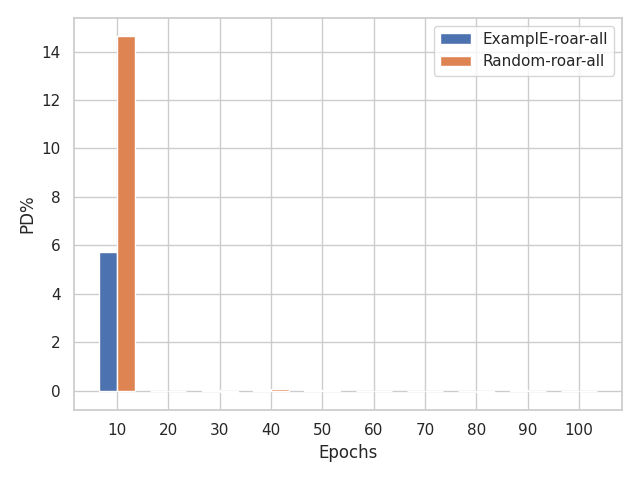}
    \end{subfigure}
    \caption{Probability differences between ExamplE and random baseline, TransE trained on Fb15k-237.}
\end{figure}

\begin{table}[!ht]
    \centering
    \begin{tabular}{p{12mm}|lp{10mm}p{10mm}lp{15mm}}
    \hline
       \textbf{epoch} & \textbf{average} &\multicolumn{2}{c}{rev-ROAR [\%]} & \multicolumn{2}{c}{ROAR [\%]} 
       \\  & & 1 & all & 1 & all \\
        \hline
        10 ours & -0.507 & 69.694 & 63.971 & \textbf{0.106} & 5.717 \\ 
        \hspace{0.15 in}   rand. & -0.478 & 73.757 & 62.186 & 0.013 & \textbf{14.647} \\ \hline
        20 ours & -0.066 & 25.457 & 23.819 & -0.0 & -0.009 \\ 
        \hspace{0.15 in}   rand.  & 0.007 & 27.968 & 21.456 & 0.001 & -0.011 \\ \hline
        30 ours& -0.181 & 10.911 & 8.27 & -0.003 & 0.013 \\ 
       \hspace{0.15 in}   rand.   & -0.348 & 11.117 & 8.388 & -0.001 & -0.073 \\ \hline
        40 ours & 0.009 & 5.312 & 4.735 & 0.0 & 0.034 \\ 
        \hspace{0.15 in}   rand.  & 0.096 & 5.435 & 4.154 & 0.002 & 0.048 \\ \hline
        50 ours & -0.049 & 2.707 & 2.227 & 0.03 & 0.036 \\ 
        \hspace{0.15 in}   rand.  & 0.021 & 2.863 & 2.196 & 0.031 & 0.022 \\ \hline
        60 ours & -0.112 & 1.635 & 1.456 & -0.002 & -0.016 \\ 
        \hspace{0.15 in}   rand.  & -0.057 & 1.718 & 1.332 & -0.001 & -0.007 \\ \hline
        70 ours & -0.135 & 0.894 & 0.786 & -0.0 & -0.011 \\ 
        \hspace{0.15 in}   rand.  & -0.072 & 0.94 & 0.721 & -0.0 & -0.015 \\ \hline
        80 ours & 0.053 & 0.574 & 0.496 & -0.001 & -0.001 \\ 
        \hspace{0.15 in}   rand.  & 0.002 & 0.599 & 0.456 & -0.001 & -0.038 \\ \hline
        90 ours & -0.153 & 0.384 & 0.325 & -0.0 & 0.028 \\ 
        \hspace{0.15 in}   rand.  & -0.421 & 0.402 & 0.303 & -0.0 & -0.0 \\ \hline
        100 ours & -0.153 & 0.23 & \textbf{0.203} & -0.007 & -0.01 \\ 
        \hspace{0.15 in}   rand.  & -0.03 & \textbf{0.244} & 0.176 & -0.007 & -0.005 \\ \hline
    \end{tabular}
    \caption{TransE on Fb15k-237 - Probability difference between original model and models retrained using two different scenarios ROAR and rev-ROAR considering most influential triple (1) and all triples from the obtained explanation (all). We can see that when retraining the model with only a single triple of given predicate (rev-ROAR-1) the model can recover from it's initial almost 70\% probability drop at epoch 10th to a little over 0.2 difference at epoch 100th, we can observe the same pattern but faster for the training with full explanation. On the other hand when we look at the ROAR experiment we can see that removing a single triple has only influence epoch 10th of training with 0.1\% probability drop, this is increased when all triples are removed to ~6\%.}    
\end{table}

\begin{table}[!ht]
    \centering
\begin{tabular}{p{12mm}|lp{10mm}p{10mm}lp{15mm}}
    \hline
       \textbf{epoch} & \textbf{average} &\multicolumn{2}{c}{rev-ROAR [\%]} & \multicolumn{2}{c}{ROAR [\%]} 
       \\ & & 1 & all & 1 & all \\
        \hline
        10 ours & 0.0 & 13.515 & 15.018 & \textbf{1.118} & \textbf{1.696} \\ \hline
        \hspace{0.15 in}   rand. & -0.004 & - & - & 0.089 & -0.306 \\ \hline
        20 ours& -0.004 & 26.975 & 29.598 & 0.008 & 3.5 \\ \hline
        \hspace{0.15 in}   rand. & -0.027 & - & 25.585 & -0.021 & -0.058 \\ \hline
        30 ours& 0.018 & 34.791 & 37.518 & 0.031 & 1.198 \\ \hline
        \hspace{0.15 in}   rand. & 0.005 & 34.384 & - & -0.046 & 0.035 \\ \hline
        40 ours& -0.004 & 38.778 & 42.265 & 0.048 & 0.041 \\ \hline
        \hspace{0.15 in}   rand. & 0.02 & - & - & 0.021 & 0.036 \\ \hline
        50 ours & 0.006 & 40.661 & 44.588 & 0.038 & 0.095 \\ \hline
        \hspace{0.15 in}   rand. & 0.025 & 42.159 & 41.134 & -0.01 & 0.004 \\ \hline
        60 ours& 0.015 & 42.071 & 45.949 & 0.054 & 0.018 \\ \hline
        \hspace{0.15 in}   rand. & 0.042 & 43.586 & 42.388 & 0.011 & 0.028 \\ \hline
        70 ours& -0.032 & 43.149 & 46.509 & 0.071 & 0.021 \\ \hline
        \hspace{0.15 in}   rand. & -0.309 & 44.506 & - & 0.005 & -0.002 \\ \hline
        80 ours& -0.017 & 42.835 & 46.048 & 0.025 & 0.017 \\ \hline
        \hspace{0.15 in}   rand. & -0.252 & 44.057 & - & 0.001 & -0.02 \\ \hline
        90 ours& 0.047 & 42.512 & 45.655 & 0.01 & 0.012 \\ \hline
        \hspace{0.15 in}   rand. & 0.058 & 43.58 & - & 0.011 & 0.013 \\ \hline
        100 ours& -0.008 & 41.778 & \textbf{44.503} & 0.023 & 0.002 \\ \hline
        \hspace{0.15 in}   rand. & 0.01 & \textbf{42.673} & 41.57 & 0.021 & 0.014 \\ \hline
    \end{tabular}
       \caption{TransE on WN18RR - Probability difference between original model and models retrained using two different scenarios ROAR and rev-ROAR considering most influential triple (1) and all triples from the obtained explanation (all). We can see that when retraining the model with only a single triple of given predicate (rev-ROAR-1) the model cannot recover from initial ~14\% probability drop at epoch 10th instead it worsen to reach it's peak at around epoch 70th (amounting to ~43\%) to settle on nearly 42\% difference at epoch 100th, we can observe the same pattern but faster for the training with full explanation. On the other hand when we look at the ROAR experiment we can see that removing a single triple has only influence epoch 10th of training with 1.1\% probability drop, this is increased when all triples are removed to ~1.7\%, the difference above epoch 10th is smaller than 1\%.}    
\end{table}

\begingroup
\begin{table*}[htb!]
\begin{center}
\begin{tabular}{p{20mm}|lp{10mm}p{10mm}lp{10mm}p{10mm}p{10mm}p{10mm}p{10mm}p{10mm}}
\toprule
    Approach &  \multicolumn{2}{c}{TransE} & \multicolumn{2}{c}{ComplEx}& \multicolumn{2}{c}{DistMult} & \multicolumn{2}{c}{ConvE} \\

     & H@1 & MRR
    & H@1 & MRR
    & H@1 & MRR
    & H@1 & MRR \\
    \midrule \\
Fb15k-237   & 0.20    & 0.30    &   0.21  & 0.31 & 0.21 & 0.30 & 0.21 & 0.30 \\
    \hline \\
WN18RR &  0.05   & 0.22    &   0.47  & 0.50 & 0.43 & 0.47 & 0.44 &  0.47\\

\end{tabular}
\caption {\label{tab:fb15k237_results}MRR and Hits@1 on the Fb15k-237 and WN18RR benchmark datasets.} 
\end{center}
\end{table*}
\endgroup

\begin{table}[!ht]
    \centering
    \begin{tabular}{l|ll}
    \hline
        dataset & total time [s] & time/triple [s]  \\ \hline
        FB15k-237 & 104069.3 & 5.09 \\ 
        CODEX & 60364.4 & 5.85  \\
        WN18RR & 36048.5 & 12.33  
    \end{tabular}
    \caption{Time to obtain explanations for triples for respective TransE models trained on different datasets, total time for all triples in the test dataset, time per triple. This also include cases when no explanations where found. We also generated explanations for ComplEx model for FB15k-237 and the respective times were 196682.1 for all test triples and 9.62s per triple - almost twice as much as for a TransE model, which make sense as ComplEx model due to its architecture has twice as much embeddings. For WN18RR the times for ComplEx model where respectively: 38289.0s and 13.09s.}
\end{table}

\begin{figure}
    \centering
\includegraphics[width=0.5\textwidth]{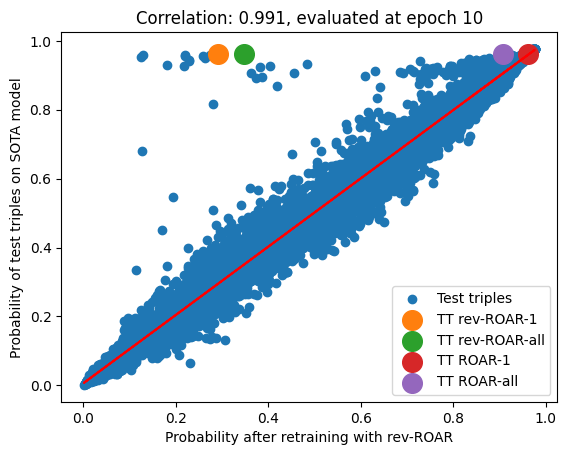}
    \caption{TransE on Fb15k-237 - Probabilities correlation before and after retraining model with reversed ROAR only explanations is left in the training dataset among triples with same predicates as target triple. Above epoch 20 the Pearson correlation coefficient is 1 and predictions are perfectly correlated.}
    \label{fig:my_label}
\end{figure}

\begin{figure}
    \centering
    \includegraphics[width=0.5\textwidth]{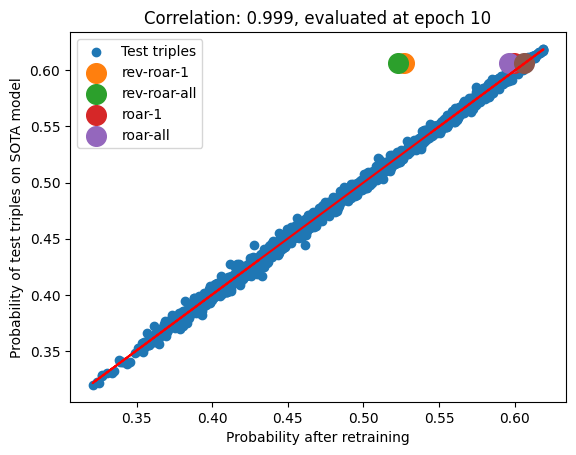}
   \caption{TransE on WN18RR - Probabilities correlation before and after retraining model with  ROAR.}
    \label{fig:my_label}
\end{figure}

\begin{figure}
    \centering
    \includegraphics[width=0.5\textwidth]{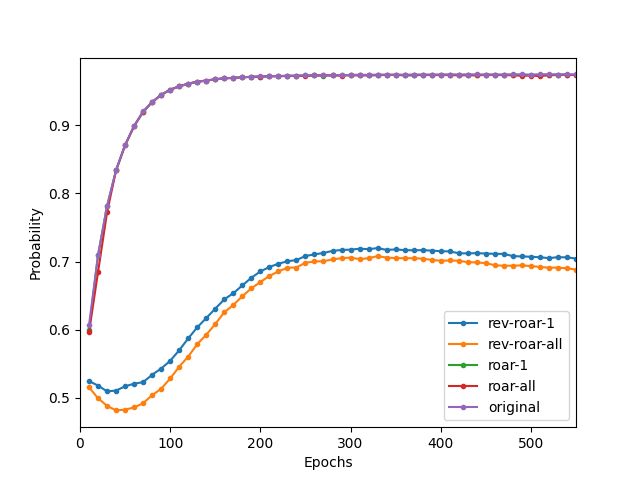}
    \caption{TransE on WN18RR. Target triple probability across different epochs.}
    \label{fig:my_label}
\end{figure}

\begin{figure}
    \centering
    \includegraphics[width=0.5\textwidth]{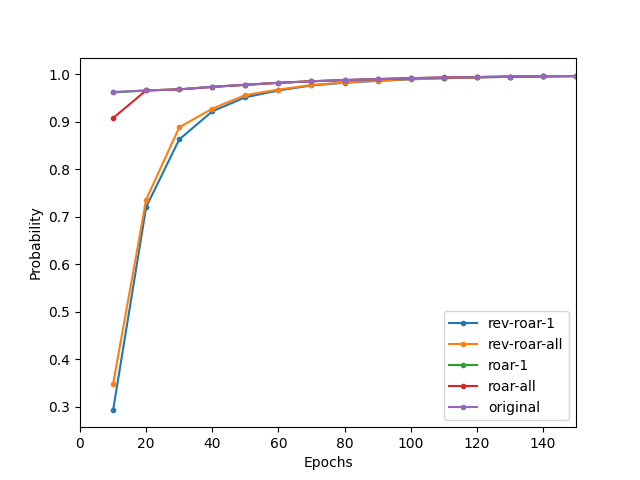}
    \caption{TransE on Fb15k-237. Target triple probability across different epochs.}
    \label{fig:my_label}
\end{figure}

\section{Related Work}\label{sec:related_work}
The most basic way to identify influential triples would be to perform a simple search over all possible triples that could be removed from the dataset and perform retraining after each such modification of the dataset. This approach is very inefficient as it requires many retrainings of the model. For example if explanation size we are interested in is equal to $|e| = 1$ we need $n$ retrainings of the model for each triple, when $n$ is the number of triples in the training dataset. The number of retrainings is increasing if we allow the explanation to be greater than 1, $|e| > 1$.

This section contains related work and up-to-date SOTA. It starts with brief introduction. Following, each work is described with differentiation factor for ExamplE in mind. Finally, comparable dimensions across presented works conclude the section. Basic principle of the majority of explanation methods presented below is as follows: they try to identify such existing links in the graph that their removal will strongly decrease the probability of the predicted link (this holds for ExplaiNE \cite{kang_explaine_2019} and GNNExplainer \cite{ying_gnnexplainer_2019} but not for GraphLIME \cite{huang_graphlime_2020}).Worth mentioning is that none of these studies conducted user-studies on effectiveness of human-readability of provided explanations or even whether the list of links/subgraph is enough to constitute the explanation.

Apart from work on explainability aspects of Knowledge Graph Embedding models we would like to bring attention to a similar but seemingly different subject of robustness and adversarial attack approaches for Knowledge Graph Embedding models. In recent work, Bhardwaj et al \cite{bhardwaj_poisoning_2021} explored methods of poisoning KGEs with relation inference patterns, which aims at targeting influential triples and design attacks based on it. Another work by Betz et al \cite{betz_adversarial_2022} introduced adversarial explanations where they identify regularities in the knowledge graph and plan attacks based on them.

In \cite{pezeshkpour_investigating_2019}, authors investigated robustness of knowledge graph embedding models with regards to removal or addition of an influential triple to the training set.

\paragraph{ExplainE} \cite{kang_explaine_2019} - counterfactual explanations for network embeddings based link prediction. As an explanation it identifies a set of existing links that influenced the predicted link. It quantifies how a predicted link would be affected by weakening an existing link. It provides a mathematical formula for choosing the links for small networks and an approximation for large(r). The method ranks most dominant links from among the 1-hop neighbourhood of the node that the link leads to and refers to it as an explanation for the predicted link to this node.
Scalability, the approximate method scales to bigger graphs but experiments reported in the paper were conducted on datasets with up to 66,597 links with small embedding sizes up to 16 (here we consider embedding sizes of 350+) and dataset of size starting from 80,894 number of links. The article does not provide guidelines on how to choose parameters. Each graph structure requires derivation of mathematical formulas for obtaining explanations, in the paper the author presented a detailed approach for CNE network and general guidelines for other networks. The code was not made public, reproducing results not possible.
The method presented here is not derived from mathematical formulas based on network structure; it is a heuristic method that has some common parts with ExplainE. First one is that both methods return as an explanation a ranked list of links from 1-hop neighbourhood yielded as important for the given predicted link. One difference here is that ExplainE considers only the neighbourhood of the node that the predicted link leads to while it considers the neighbourhood of both nodes that the predicted link connects. Another thing is the form of explanation, provided along the relevant links from 1-hop neighbourhood, also supporting examples of triples that have similar characteristics as in example-based explanation returning respective subgraphs.

\begingroup
\begin{table*}[htb!]

\setlength\heavyrulewidth{0.45ex}
\begin{tabular}{p{20mm}|lp{20mm}p{40mm}lp{12mm}p{15mm}p{15mm}p{8mm}}
Method            & Task & Modality/ GML Family & Type of XAI method & code & Format & Special Training Mode & Memory Footprint & Speed\\
\toprule
Gradient Rollback    & LP* & KG+KGE & Gradient-based Attribution      & + & T* & + & $\uparrow$ & $\downarrow$    \\
ExplainE              & LP & KG+KGE & Counterfactuals & -   & T  & + & - & -                       \\
\midrule
\gray{GNNExplainer}        & \gray{NC*}  & \gray{GNN} &  \gray{Perturbation-based method}       & \gray{+} & \gray{T}, \gray{NF} & \gray{-} & \gray{-} & \gray{-         } \\
\gray{GraphLIME}            & \gray{NC}   & \gray{GNN} & \gray{Surrogate method} & \gray{+} & \gray{NF} & \gray{-} & \gray{-} & \gray{-} \\
\gray{ZORRO} & \gray{LP}             & \gray{GNN} & \gray{Perturbation-based method, Instance-wise feature selection (IFS)} & \gray{planned} & \gray{T, NF}  & \gray{-} & \gray{-} & \gray{-}\\
\midrule
\textbf{ExamplE} & LP   & KG+KGE   & Example-based explanation & planned & T, Meta & - & $\downarrow$ & $\uparrow$
\end{tabular}

\caption {\label{tab:table4} Comparison of different explainability methods designed for relational learning models. Note all models are post-hoc methods, i.e. methods that explain pre-trained, black-box, opaque neural predictors. Being instance-level means that each method explains a specific prediction, instead of providing a global interpretable explanation for the entire predictive pipeline. NC* - Node Classification, LP* - Link Prediction, GNN- Graph Neural Network, T* - Triples, NF - Node Features, Meta - Metadata.}

\end{table*}
\endgroup

\paragraph{GNNExplainer} \cite{ying_gnnexplainer_2019} - identifies a subgraph and a subset of nodes that is relevant for the prediction and is designed for Graph Neural Network models (another family of relational learning models). For this reason, we consider only GNNExplainer’s output explanation, since the logic behind the training model is different from the knowledge graph embedding case. GNNExplainer is based on the  optimization of mutual information between distribution of possible subgraph structures and the predictions. It provides an explanation for an instance or for a group of samples by aggregating them and extracting a prototype. GNNExplainer optimises for mutual information in order to learn two masks: a graph mask that selects an important subgraph and a feature mask of unimportant node features, which is later used to reduce the input graph to get a minimal explanation subgraph which can be drawn from the multi-hop neighbourhood. GNNExplainer goes beyond the 1-hop neighbourhood to construct the explanation subgraph, making it applicable for more complex dependencies in the neighbourhood of the considered nodes. On the other hand it is not applicable to knowledge graph embeddings as is, and it is not an example-based approach. Substantial analysis is needed on its applicability to KGE models.

\paragraph{GraphLIME} \cite{huang_graphlime_2020} adapts the popular interpretability method LIME to KGE models. It also introduces a non-linearity in the method. This approach however only highlights important fragments of knowledge embedding vectors, and does not link back to the original graph making it not interpretable to human users (knowledge graph embedding dimensions are not directly interpretable).

\paragraph{GradientRollback} \cite{lawrence_explaining_2020} is a method to explain knowledge graph embedding models for link predictions. It works by storing gradient updates in a separate influence matrix per every training example T (during training) and also per every unique entity and relation in a triple. It then refers to this gradient update matrix during the explanation phase when updates regarding one training triple (T) are subtracted from the parameters matrix to obtain a new  parameter matrix that simulates the situations when we retrain the model without T. This approach requires enabling a special training mode with batch size of 1 and requires much more memory than the initial dataset size to store training artefacts.
E.g. For embedding size of k=100 and training dataset of size 272115 (Fb15k-237) apart from model parameters during training we are storing around 272115 * 3 * 100 (may be smaller if self relations are present, in such case we only have one entity per triple, and 3 comes from 3 elements of triple 2 entities and relation). For Fb15k-237 the size of this extra matrix is equal to 2.4G. For simulation of the size of the training artefacts with respect to the dataset size for GradientRollback, please see appendix Figure 12. This method although time and memory consuming traces all steps in the training leading to probably more accurate results.

ZORRO \cite{funke_zorro_2022} is designed for graph neural networks. It provides explanations via masking that decreases the prediction error between the model trained on the full dataset and an explanation.

\section{Conclusion}\label{sec:conclusion}

In this work we introduce a novel format of explanations: explanation graph and ExamplE heuristics to generate explanation graphs and time-efficient batch mode for generating influential examples only. Evaluation protocol involving novel XAI test set for evaluating interpretability of explanations for the users. We will make the code available along with the dataset as a part of a new release of AmpliGraph Python library. For future work we would like to compare how this approach work on the GNN architectures since it is a model agnostic heuristic. One disadventage that we are aware of is that our approach is a heuristics, we are doing some follow-up research on how to provide estimation guarantees for this approach and if it is even possible. But every coin has two sides, on the other side we have decided to go forward with the heuristic approach because of immense memory consumption and slow execution time of other methods. The question remains is heuristic good enough or should we look into exact methods?

\section{Acknowledgements}\label{sec:ack}
The work was partially funded by the European Union (EU) Horizon 2020 research and innovation program through the Clarify project (875160).

\bibliographystyle{aaai}
\bibliography{bibliography}

\section*{Appendix}
\subsection{A Computational Complexity Analysis}
We have analyzed computational complexity of the ExamplE in the scenario of batch explanations.
ExamplE requires access to the training dataset so space complexity starts from $O(t)$.
Given:
$e$ - number of entities (e.g.: 14,541 in Fb15k-237).
$k$ - embedding vector dimension (e.g. k=400, TransE on Fb15k-237).
$m$ - number of nearest neighbours considered (parameter of ExamplE default m=25).
$t$ - number of triples in the train set.
$x$ - number of examples, as explanation $x << t$.
we can split computational complexity into steps:
1) Sampling - this step is entirely dependent on the nearest neighbour algorithm implementation, in the experiments we used implementation provided in sklearn, which by default tries to adjust parameters for best efficiency. In the worst case scenario it uses a brute force approach which complexity of the prediction time is $O(e \times k \times m)$ with negligible complexity of initialization of the algorithm and negligible space complexity too. In the best case scenario kNN algorithm tries to adjust the inner data structure for optimized inference time with the cost of initialization and space e.g. in the case of KD-Tree it is $O(k \times e \times log(e))$ of extra initialization time and $O(k \times e)$ space with a benefit of inference time being $O(m \times log(e))$. ExamplE needs to find m nearest neighbours for both subject and object entity (in the default case) in this step.
2) Filtering  for example triples - in this step we need to take a cartesian product of obtained sets of neighbours in step 1: which leaves us with $O(m^2)$ (default case, in full case it is $O(m^3)$ if we are considering predicates embedding as well) and forces us to filter examples according to the dataset. First we are mapping it into a tuples ($O(t)$, where $t$ is a number of train triples), than we utilize sets intersections implementation in Python with complexity of $O(min(t,m^2))$. In the post-processing step we compute the score per each example obtained. The computational complexity in batch explain is always dependent on the number of target triples to obtain explanations for.

\begin{table}[]
\begin{tabular}{lllll}
& initialization &  \\
step            &  time  & sapce         \\
$sampling_{KD-Tree}$     & $O(k \times e \times  log(e))$ & $O(k \times e)$     \\
$sampling_{brute-force}$ & $O(1)$ & $O(1)$ \\
\end{tabular}
\caption{Computational time and space complexity of initialization phase.}
\end{table}

\begin{table}[]
\begin{tabular}{lllll}
& prediction &  \\
step            &  time  & sapce         \\
$sampling_{KD-Tree}$ & $O(m \times log(e))$ &   $O(1)$   \\
$sampling_{brute-force}$ & $O(e \times k \times m)$ & $O(1)$ \\
product & $O(m^2)$  & $O(1)$ \\
mapping & $O(t)$ & $O(t)$ \\
filtering       &      $O(min(t,m^2))$   &     $O(1)$          \\
post-processing &       $O(x)$               &    $O(1)$           
\end{tabular}
\caption{Computational time and space complexity of prediction phase.}
\end{table}

\subsection{B Implementation Details}
Table \ref{tab-hyperparams} lists training parameters for models used in the experiments.
\begin{table}[h]
\begin{tabular}{lllll}
model/dataset & WN18RR & Fb15k-237 &  &  \\
TransE        &    k=350, eta=30&   k=400, eta=30 &  &  \\
\end{tabular}
\caption{Parameters used for model training, trained with early stopping for 4000 epochs using Adam optimizer with lr=0.0001, multiclass-nll loss, seed=0, regularizer L2 with lambda=0.0001}
\label{tab-hyperparams}
\end{table}

\subsection{C Models Calibration}
Knowledge Graph Embedding models predictions are uncalibrated - meaning they do not represent probability of the triple but rather a plausability score. Explainability approches presented in this work requires models to be calibrated.
Achieving such property with knowledge graph embedding models requires additional
post-processing to make sure the returned predictions can be interpreted as probabilities. We calibrate the trained model using procedure described in \cite{tabacof_probability_2020} on the validation test. Figures 9 and 10 shows the reliability diagram for uncalibrated and calibrated scores compared with perfectly calibrated reference for two datasets. We can see that uncalibrated
scores (in blue) are unevenly distributed and concentrated around the lower part of predicted values, whereas calibrated scores (in orange) although not perfect are distributed much more evenly across the mean predicted values space. By calibrating pre-trained knowledge graph embedding models to return values between 0 and 1 that are more
evenly distributed, we return more reliable and interpretable probabilities of triples.

\begin{figure}
    \centering
\includegraphics[width=0.49\textwidth]{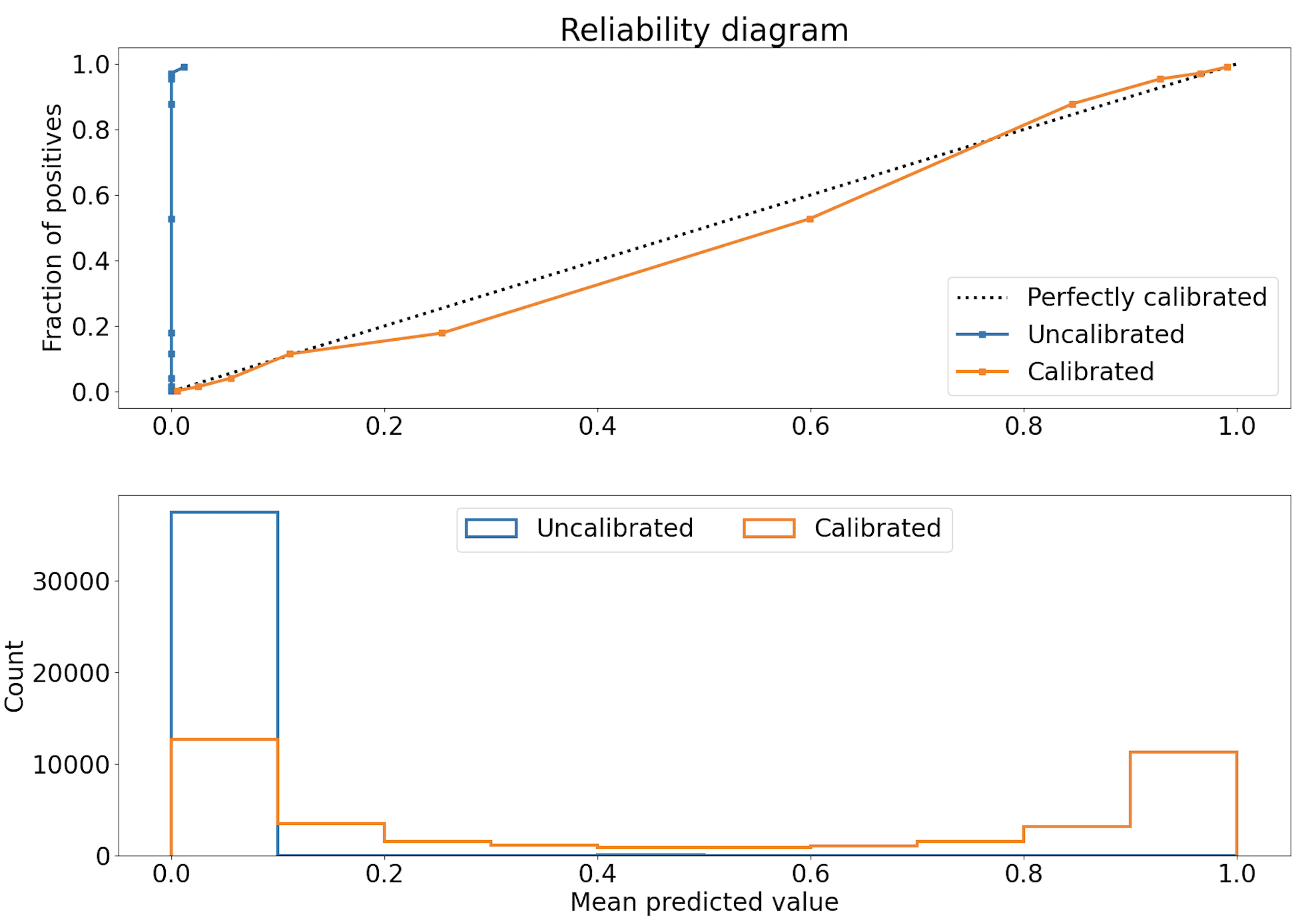}
    \caption{Reliability diagram of the TransE model's calibration for FB15k-237 dataset.}
    \label{fig:my_label}
\end{figure}
\begin{figure}
    \centering
\includegraphics[width=0.5\textwidth]{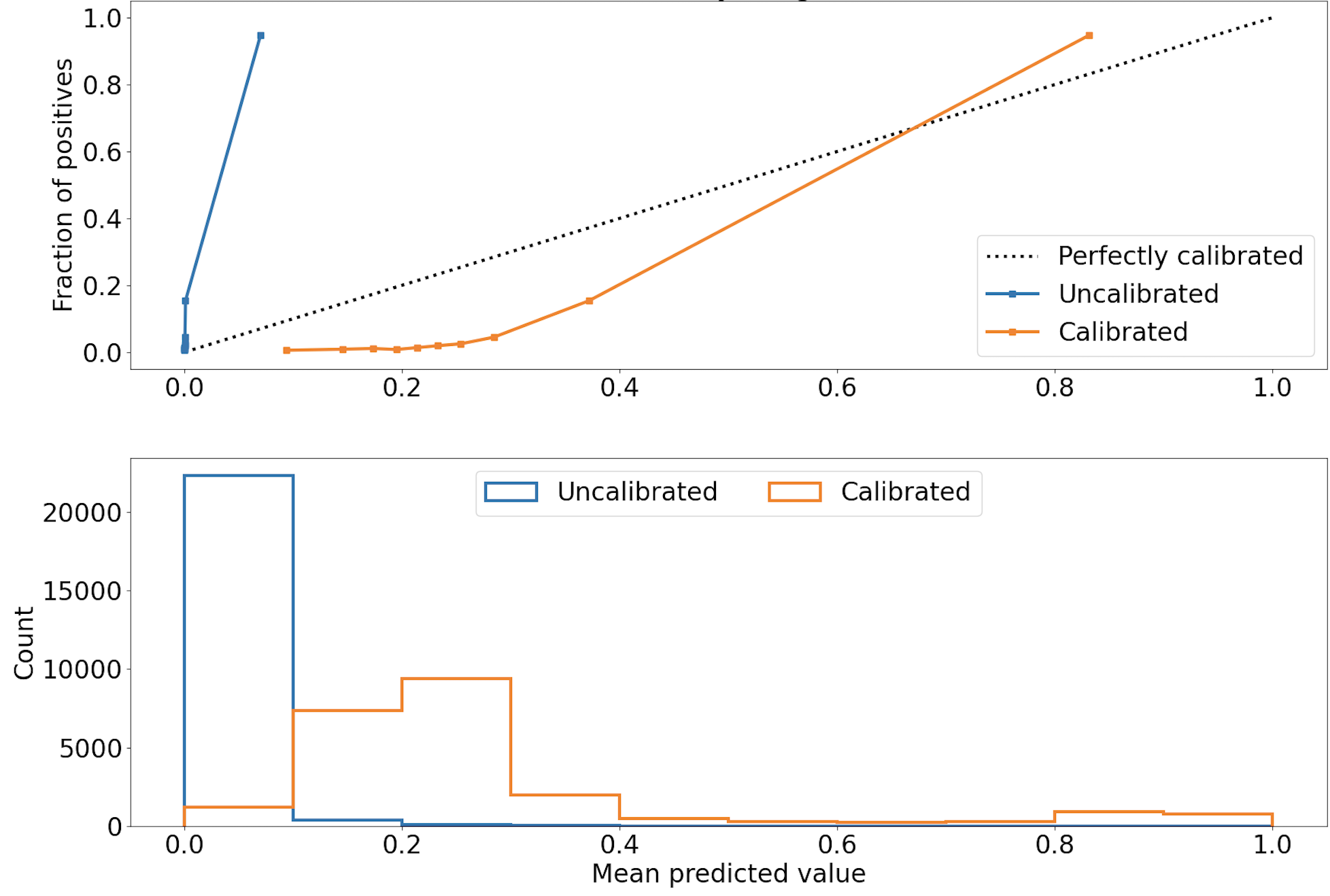}
    \caption{Reliability diagram of the TransE model's calibration for WN18RR dataset.}
    \label{fig:my_label}
\end{figure}

\subsection{E XAI Fb15k-237 Dataset construction}

At the moment there are no available and accepted benchmark datasets for human evaluation of knowledge graph link prediction explainability methods. Each paper published up-to-date follows different evaluation approach. It is, therefore, difficult to compare and draw conclusions about which of the methods gives the best results.

Such evaluation dataset should be in itself understandable for a layman. This is not true for current common benchmark datasets in the knowledge graph embedding community. Some benchmark datasets like Freebase-derived have their entities encoded which are not understandable, the mappings could be obtained but are not comprehensive and sometimes not well-aligned. Each benchmark dataset have test triples that constitute a part of the whole dataset (10-20\%), for example Freebase fb15k-237 contains 20438 test triples, which would be expensive (and not feasible) to evaluate explainability methods on with a human evaluation protocol. For a rough estimate to evaluate 100 triples it would take up to 10 minutes for a person to assess if the triple is understandable (only) or not. To assess 20438 triples this would require from one annotator to spend ~34 hours annotating (almost a work-week), a comprehensive evaluation would require more than 1 annotator to assess each test case - at least 3 (preferably 5). This is only if we consider assessing a single triple to be understandable on its own without mentioning the full explanation of why such connection was suggested, such explanation could span from 2 triples (target triple + explaining triple) to the size of a graph. Taking into account human evaluation cost and effort of such evaluation we can say this is unfeasible to achieve for every published method on every benchmark dataset. This inability to assess the usefulness of produced explanations to the audience, including lack of common framework and high cost of it, is a common obstacle that researchers working on such problems encounter and therefore refrains from running such studies limiting the evaluation to the synthetic metrics.

\begin{figure*}[!ht]
    \centering
    \includegraphics[width=\textwidth]{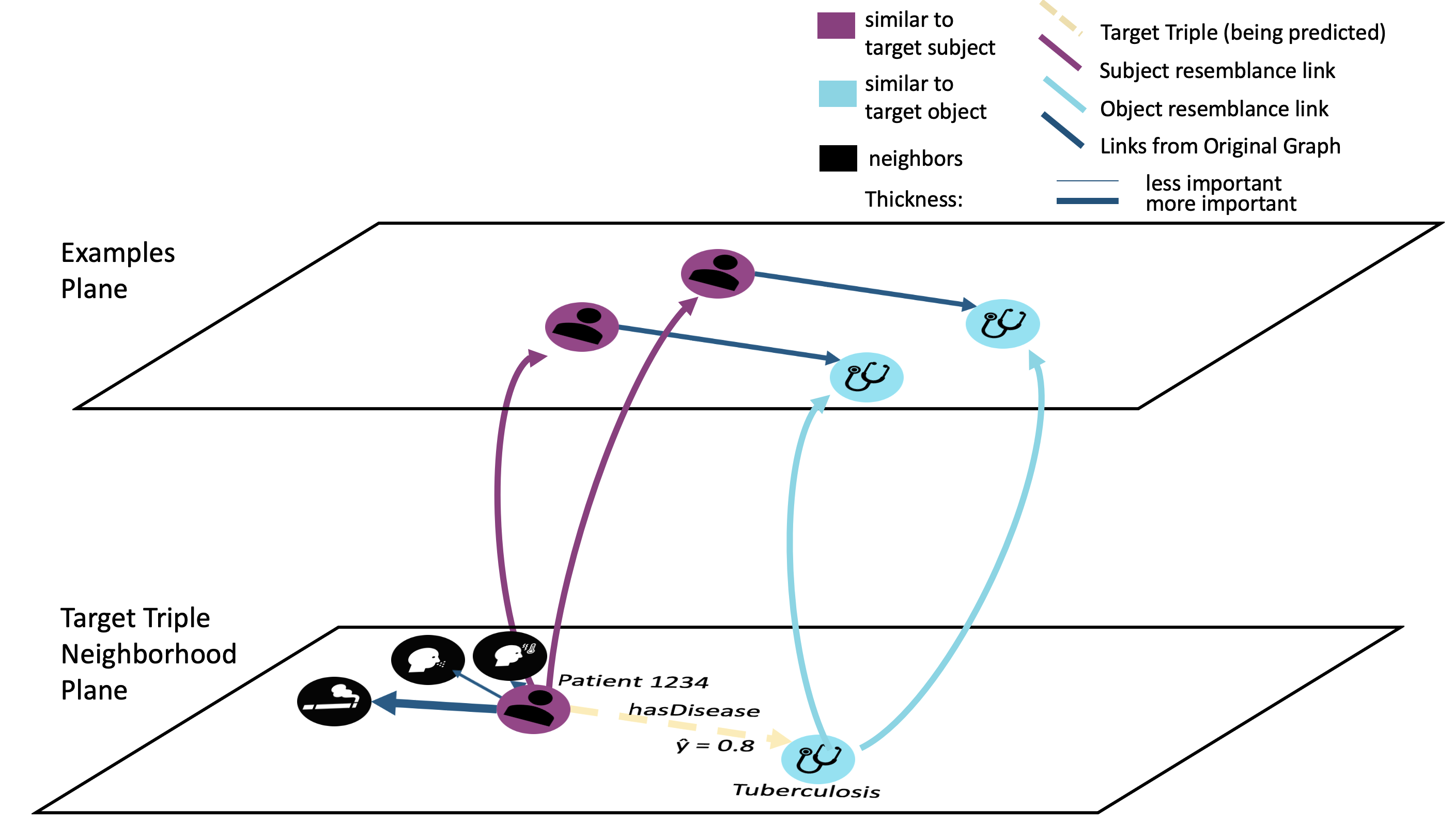}
    \caption{Explanation Graph anatomy. Explanation Graph consists of two main components: examples (shown above on an Examples Plane) and prototype (shown above in a Target Triple Neighbourhood Plane). It contains both important triples from the original knowledge graph - \textbf{influential examples} as well as links representing meta-data that enriches explanation.}
    \label{fig:diagram}
\end{figure*}
Recent publications on explainable link prediction do not provide any evaluation framework that could contribute to fair comparison of explainability approaches, whereas explainability of methods requires strict protocols involving a human-evaluation \cite{doshi-velez_towards_2017}.

A contribution to rigours evaluation could be a benchmark datasets that would make it possible to compare different explainers of link prediction models. 

The following sections describe the desiderata of such dataset and then the evaluation protocol of obtaining such dataset with an example of Fb15k-237 dataset which is a benchmark dataset commonly accepted in the community for assessing link prediction models. Sub-setting already existing benchmarking datasets vs creating another dataset from scratch has an advantage of being already familiar to the community and can eliminate the problem of running explantation methods for models that benchmark results are not known. The disadvantages of such approach is preserving biases available in the original datasets, dataset becoming obsolete (e.g. Freebase was discontinued).
\subsubsection{Desiderata}
\begin{itemize}
    \item    human readable triples: huaman-readable entities and predicates (e.g. Fb15k-237).
    \item    human understandable triples (triples should be aligned with a layman understanding).
    \item    suitable for the human evaluation (limited size according to the cognitive load).
    \item    diverse across different predicates (for knowledge graphs).
    \item    no duplicates.
    \item    no self-relations.
    \item    human-filtered.
    \item    based on a benchmarked dataset (for easier elimination of problems that arise from poor models).
    \item    publicly available.
\end{itemize}

Such dataset can provide a ground for a fair comparison across different explainability methods and interpretability of their explanations. In Figure 13 we present a small user study we ran in the lab to curate a small subset of Fb15k-237 with triples that could be used in such a user study evaluation of the explanations. Similar protocol could be employed for other suitable datasets with more interpretable entities and relations (e.g. CoDEx).

\begin{table*}[!ht]
 
    \begin{tabular}{|l|l|l|l|}
    \hline
        S & P & O & Score \\ \hline
        Billy Idol & languages spoken, written, or signed & English & TT \\ \hline
        \midrule
        Johnny Marr & languages spoken, written, or signed & English & 0.00075 \\ \hline
        Chester Bennington & languages spoken, written, or signed & English & 0.00076 \\ \hline
        Morrissey & languages spoken, written, or signed & English & 0.00077 \\ \hline
        Loreena McKennitt & languages spoken, written, or signed & English & 0.00077 \\ \hline
        Gordon Lightfoot & languages spoken, written, or signed & English & 0.00080 \\ \hline
        Alan Stivell & languages spoken, written, or signed & English & 0.00080 \\ \hline
        Robert Plant & languages spoken, written, or signed & English & 0.00080 \\ \hline
        Oleg Skripka & languages spoken, written, or signed & French & 0.00091 \\ \hline
        Alan Stivell & languages spoken, written, or signed & French & 0.00091 \\ \hline
        Oleg Skripka & languages spoken, written, or signed & Russian & 0.00097 \\ \hline
        Oleg Skripka & languages spoken, written, or signed & Ukrainian & 0.00117 \\ \hline
    \end{tabular}
    \caption{Example explanation for a test triple in CODEX-M dataset - first row represents Target Triple (TT). The lower the score, the closer is example to the Target Triple.}
\end{table*}

\begin{table*}[!ht]
   
    \begin{tabular}{|l|l|l|l|}
    \hline
        S & P & O & Score \\ \hline
        Artie Lange & /influence/influence\_node/influenced\_by & Jackie Gleason & TT \\ \hline
        George Carlin & /influence/influence\_node/influenced\_by & Danny Kaye & 0.17232 \\ \hline
        Conan O'Brien (aka Big Red) & /influence/influence\_node/influenced\_by & Danny Kaye & 0.19481 \\ \hline
        Conan O'Brien (aka Big Red) & /influence/influence\_node/influenced\_by & Steve Allen & 0.24502 \\ \hline
        Bill Maher & /influence/influence\_node/influenced\_by & Steve Allen & 0.25498 \\ \hline
    \end{tabular}
     \caption{Example explanation for a test triple in Fb15k-237 dataset - first row represents Target Triple (TT). The lower the score, the closer is example to the Target Triple.}
\end{table*}

\begin{table*}[!ht]

    \begin{tabular}{|l|l|l|l|}
    \hline
        S & P & O & Score \\ \hline
        02314321 & \_hypernym & 08102555 & TT \\ \hline
        02314001 & \_hypernym & 08102555 & 0.02788 \\ \hline
        02313495 & \_hypernym & 08102555 & 0.04839 \\ \hline
        01928360 & \_hypernym & 08102555 & 0.05155 \\ \hline
        02314717 & \_hypernym & 08102555 & 0.06077 \\ \hline
        01928737 & \_hypernym & 08102555 & 0.06294 \\ \hline
        02321759 & \_hypernym & 02316038 & 0.09046 \\ \hline
    \end{tabular}
         \caption{Example explanation for a test triple in WN18RR dataset - first row represents Target Triple (TT). The lower the score, the closer is example to the Target Triple.}
\end{table*}

\begin{figure*}
    \centering
    \includegraphics[scale=0.4]{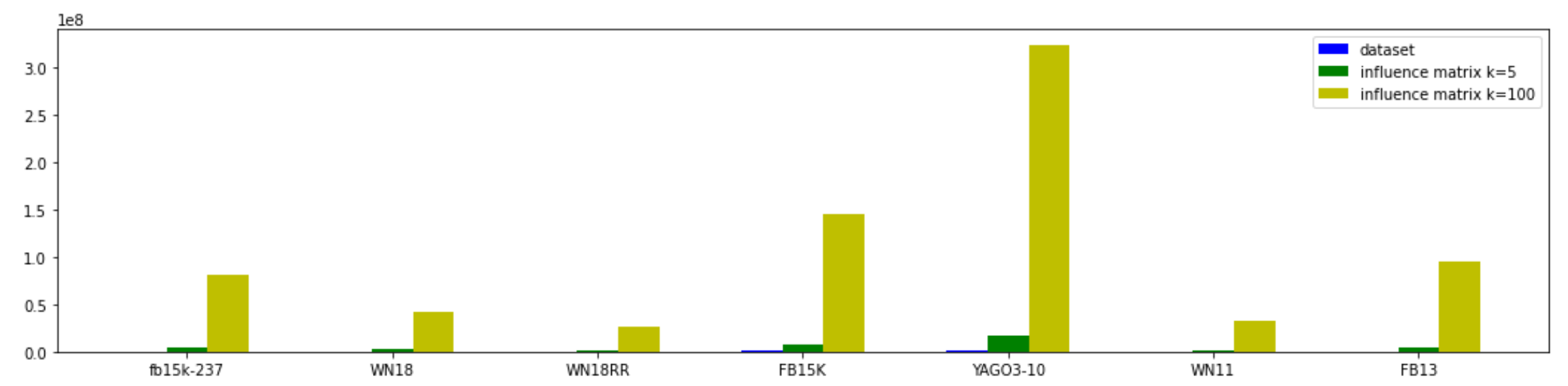}
    \caption{GradientRollback: plot with benchmark dataset sizes and estimated size of influence matrix according to increasing size of embedding (k=5 and k=100). Simulation done based on provided implementation by the authors. It does not represent exact numbers, just a rough estimation of how much space would be needed based on different embedding sizes given the method records weights changes across training.}
    \label{fig:my_label}
\end{figure*}

\begin{figure*}
    \centering
    \includegraphics[width=\textwidth]{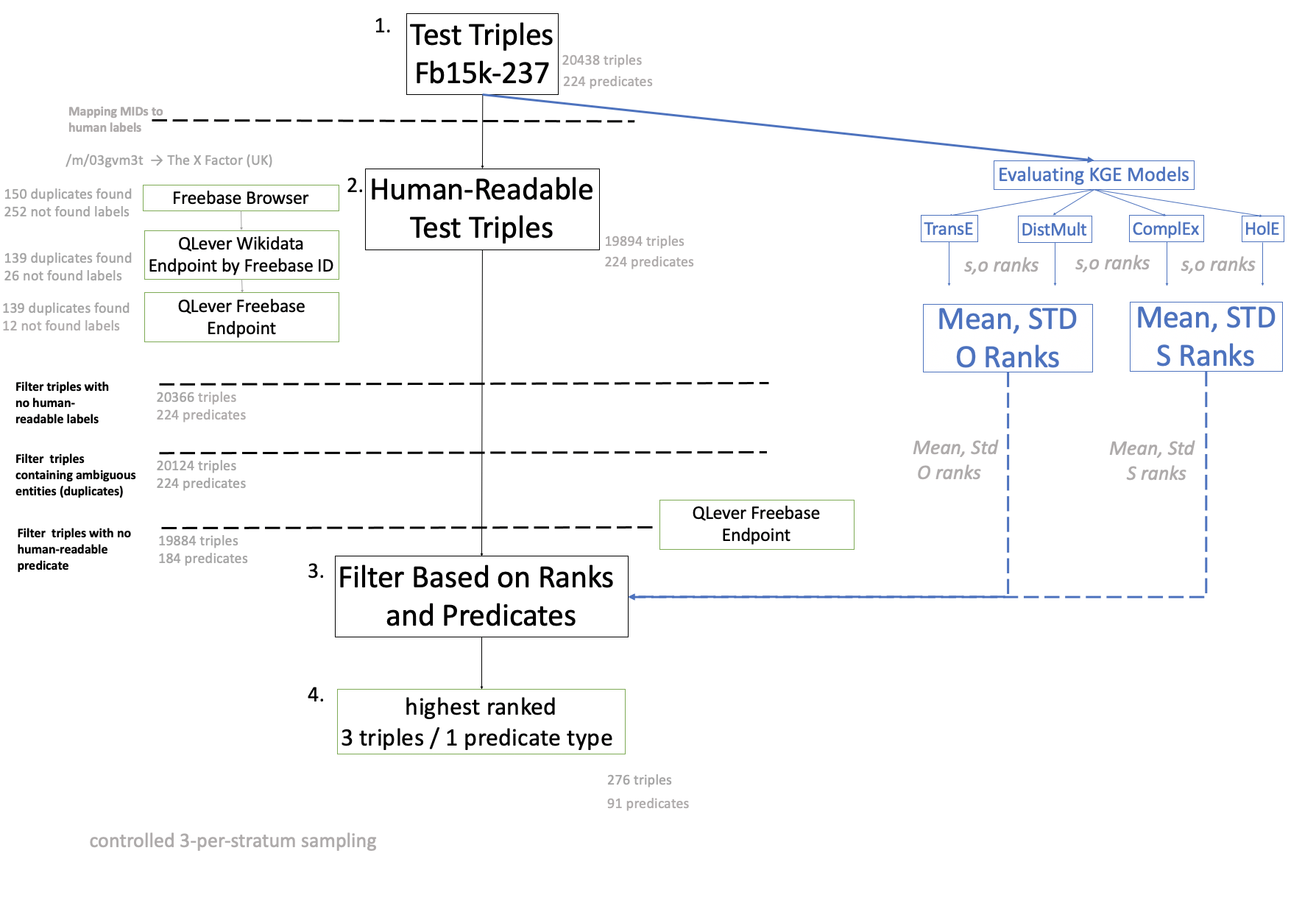}
    \caption{Validation Protocol. The diagram presents steps to be taken to get the dataset for human evaluation. It starts from the original benchmark dataset: Fb15k-237 (box number 1.) and ends at the curated subsets of triples based on properties listed before. The path highlighted in blue refers to the collection of ranks according to different models. The green box specifies selection criteria for constructing the final datasets.}
    \label{fig:val_protocol}
\end{figure*}

\end{document}